\documentclass[10pt,twocolumn,letterpaper]{article}

\usepackage{iccv}
\usepackage{times}
\usepackage{epsfig}
\usepackage{graphicx}
\usepackage{amsmath}
\usepackage{amssymb}
\usepackage{booktabs}
\usepackage{multirow}
\usepackage{wrapfig}
\usepackage{caption}
\usepackage{algorithm}
\usepackage{algorithmicx}
\usepackage{algpseudocode}
\usepackage{afterpage}
\floatname{algorithm}{Algorithm}

\usepackage[pagebackref=true,breaklinks=true,letterpaper=true,colorlinks,bookmarks=false]{hyperref}

\usepackage[breaklinks=true,bookmarks=false]{hyperref}

\iccvfinalcopy 

\ificcvfinal\fi

\begin{document}

\title{Towards General Visual-Linguistic Face Forgery Detection}
\author{
   \large Ke Sun $^{1}$, Shen Chen $^{2}$, Taiping Yao $^{2}$, Haozhe Yang $^{1}$, Xiaoshuai Sun $^{1}$, Shouhong Ding$^{2}$, Rongrong Ji$^{1}$\thanks{Corresponding author} \\
   \normalsize $^1$ Key Laboratory of Multimedia Trusted Perception and Efficient Computing, \\
   \normalsize Ministry of Education of China, Xiamen University, China \\
   \normalsize $^2$ Youtu Lab, Tencent, China
}
\maketitle
\ificcvfinal\thispagestyle{empty}\fi

\begin{abstract}
   Deepfakes are realistic face manipulations that can pose serious threats to security, privacy, and trust.
Existing methods mostly treat this task as binary classification, which uses digital labels or mask signals to train the detection model. We argue that such supervisions lack semantic information and interpretability.
To address this issues, in this paper, we propose a novel paradigm named
Visual-Linguistic Face Forgery Detection(VLFFD), which uses fine-grained sentence-level prompts as the annotation.
Since text annotations are not available in current deepfakes datasets, VLFFD first generates the mixed forgery image with corresponding
fine-grained prompts via Prompt Forgery Image Generator
(PFIG). Then, the fine-grained mixed data and coarse-grained original data and is jointly trained with the Coarse-and-Fine Co-training framework (C2F), enabling the model to gain more generalization and interpretability. The experiments show the proposed method improves the existing detection models on several challenging benchmarks. Furthermore, we have integrated our method with multimodal large models, achieving noteworthy results that demonstrate the potential of our approach. This integration not only enhances the performance of our VLFFD paradigm but also underscores the versatility and adaptability of our method when combined with advanced multimodal technologies, highlighting its potential in tackling the evolving challenges of deepfake detection.

\end{abstract}

\section{Introduction}
\label{sec:intro}
Recently, face forgery methods have achieved significant success with the growth of computer vision techniques. Such methods can manipulate facial attributes~\cite{gonzalez2018facial}, expressions~\cite{liu2019stgan}, and even identity~\cite{korshunov2018deepfakes} with high quality, which can be easily abused by malicious users and further cause severe trust issues or societal problems~\cite{tolosana2020deepfakes}. Thus, to address these issues, it is significant to develop the face forgery detection model to identify authenticity. 
\begin{figure}[t]
    \begin{center}
       \includegraphics[width=1\linewidth]{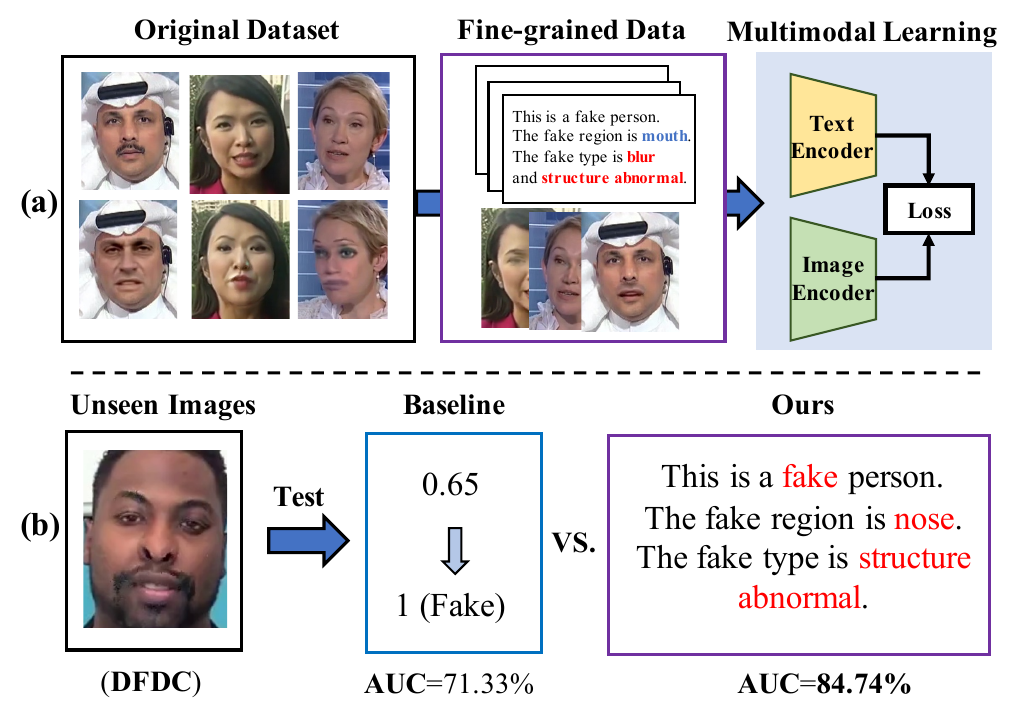}
    \end{center}
       \caption{Paradigm of our VLFFD. Traditional method trains a unimodal encoder via digitized binary labels and can only output the probability of real or fake during test time. Our method trained multimodal encoders
       with generated mixed forgery image and the fine-grained language-level annotation and can output the similarity score between the visual and sentence, which is more interpretable. Furthermore, the performance of our method outperforms the baseline $13\%$ under the unseen test data in terms of AUC.
       (Best viewed in color.)
       }
    \label{fig:intro}
\end{figure}

Since real-world forgeries often exhibit a significant domain gap from the training data~\cite{sun2021dual}, resulting in a significant drop in the performance of the detector. Consequently, the focus has shifted towards \textit{general face forgery detection}\cite{sun2021domain} to better handle unseen forgeries. 
Major solutions can be divided into \textit{forgery simulation} and \textit{framework engineering}. The former manually produces forgery samples to enhance the generalization forgery features, such as boundary~\cite{li2020face,shiohara2022detecting} and self-inconsistency~\cite{zhao2021learning}. The latter is devoted to specially designed frameworks that boost the generalization of backbones~\cite{zhao2021multi,sun2021dual}.
However, the aforementioned methods are basically trained with binary labels under a unimodal setting on the ImageNet~\cite{deng2009imagenet} pretrain model. 
Such a coarse-grained paradigm lacks fine-grained guidance and fundamental semantic information, making the model easy to overfit on the specific forgery clues in the training dataset, leading to sub-optimal performance when transferred to unseen forgery types.
Furthermore, only outputting a single confidence of real or fake lacks interpretability, especially under some strict situations such as forensics. 

Recently, multi-modality Visual Language Pretrain (VLP) models, as represented by the CLIP~\cite{radford2021learning}, have the ability to align semantic information hidden in natural language with the visual signal and
have demonstrated impressive zero-shot capabilities and robust generalization across various applications. 
Thus, we raise the following question: \textit{Could this multimodal paradigm be harnessed to extract more comprehensive and understandable information within the realm of generalized face forgery detection?}

To this end, we first design a paradigm that leverages a pre-trained multimodal model CLIP, integrating natural language to provide fine-grained supervision to improve the generalization and interoperability for face forgery detection. We call this new paradigm Visual-Linguistic Face Forgery Detection VLFFD. The difference between the traditional method and ours is shown in Fig.~\ref{fig:intro}. Our approach enhances the original dataset by enriching it with fine-grained data and textual prompts, followed by training within a multimodal learning framework. In the testing phase, our method not only achieves superior generalization but also offers detailed insights.
Specifically, we use the CLIP model as the initial feature extraction and first generate mixed forgery images with fine-grained annotations by disentangling and taxonomizing
the dataset of real and fake pairs via quantitative criterion, named Prompt Forgery Image Generator (PFIG).
Such annotations can further form the sentence-level prompts, which can provide fine-grained semantic annotations for the mixed forgery image. 
Subsequently, we
design a new contrastive-based multimodal mechanism to jointly train the coarse-grained original data and fine-grained mixed forgery data in a visual-linguistic manner, named Coarse-and-Fine Co-training framework (C2F). 
Compared with the traditional framework, the proposed VLFFD has three main advantages: 1) semantic information hidden in the CLIP can be exploited in the general face forgery detection task. 2) the fine-grained prompt can provide more precise guidance to encourage the image encoder to extract generalized forgery clues. 3)  traditional methods can only output the confidence of real or fake, while our framework can indicate the sentence about the forgery regions and types, enabling us to understand the basis of the discrimination and have more interpretability.

Additionally, our empirical evidence showcases that the fine-grained semantic annotations generated by our PFIG module can be utilized to fine-tune multimodal Large Language Models (LLM). This further highlights the potential and versatility of our proposed method.

Our main contributions can be summarized as follows:

\begin{itemize}

    \item We are the first to introduce the Visual Language Pretrain model and multimodal learning paradigm into the general face forgery detection task.
    
    \item We propose a pipeline to generate the mixed forgery image with fine-grained annotation within the existing dataset.
    
    \item We propose a Coarse-and-Fine Co-training framework that improves generalization and interpretability by training with forgery data of varying granularity.
    
    \item Extensive experiments and visualizations demonstrate the effectiveness of our method against SOTA competitors.
    
\end{itemize}


\section{Related Work}
\label{sec:related}
\subsection{General Face Forgery Detection}
Since the performance of the traditional face forgery detection method drops significantly when tested on unseen attacks, some works are devoted to relieving the generalizing problem from architecture engineering or forgery simulation. The former manually produces forgery samples to enhance the generalization forgery features that are easily overlooked by networks, such as SLADD\cite{chen2022self} and SBI~\cite{shiohara2022detecting}. The latter works are devoted to a specially designed framework that can boost the generalization of backbones. 
For example, GFF~\cite{sun2021dual} uses high-frequency stream to relieve the textures bias, and DCL~\cite{sun2021dual} leverage a supervised contrastive learning framework to preserve the variance among forgery instances. Another line of work introduces modalities other than RGB images to mine for forgery traces. For example, some works leverage frequency information as an auxiliary modal to obtain subtle forgery clues, such as DCT~\cite{qian2020thinking}, spatial-phase~\cite{liu2021spatial}, and SRM~\cite{luo2021generalizing}.
Other works exploit the motion of lips as additional supervisory signals~\cite{haliassos2021lips,haliassos2022leveraging}. Besides, some methods also leverage mask as auxiliary supervision signal\cite{chen2021local,stehouwer2019detection}.
However, these methods ignore the fine-grained semantic information, which can help the model obtain more generalization features.
\begin{figure*}[!t]
    \begin{center}
    
       \includegraphics[width=0.95\linewidth]{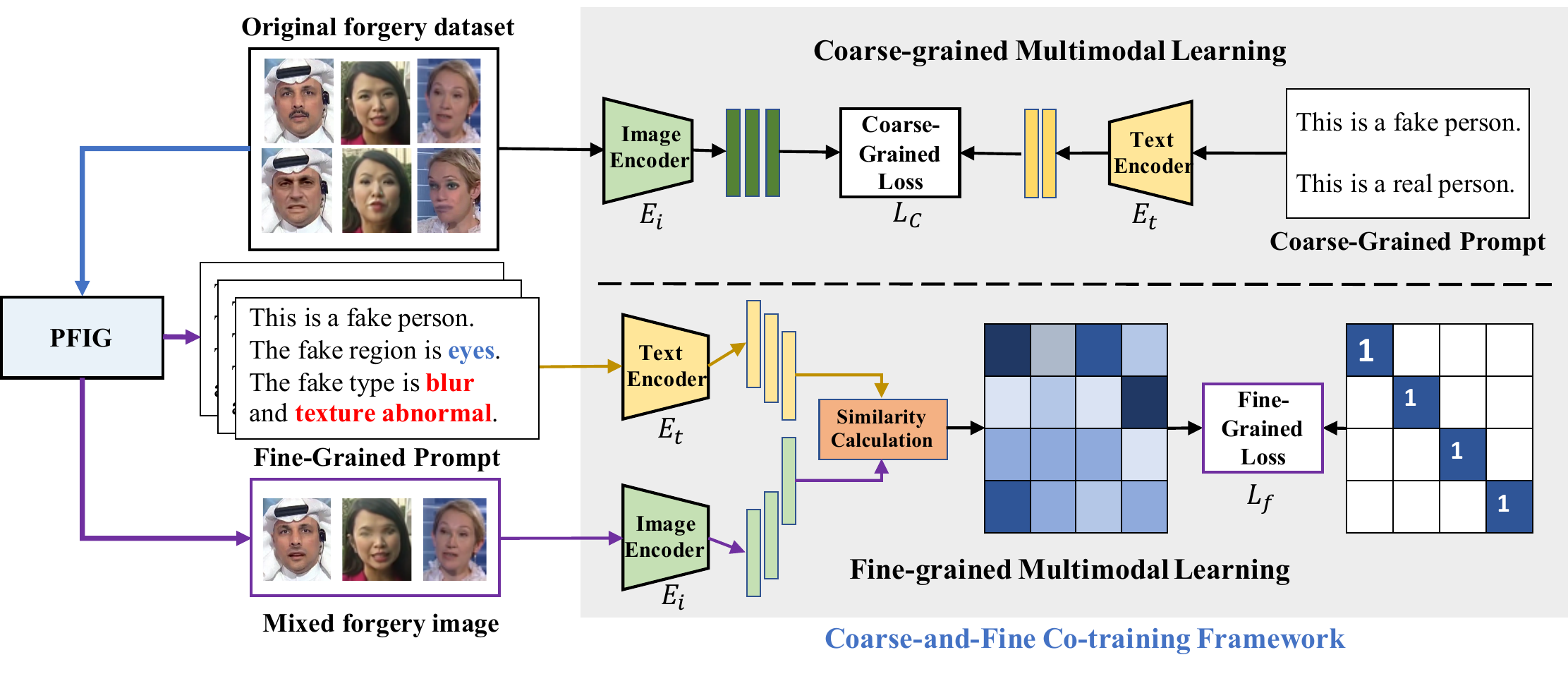}
    \end{center}
       \caption{The overview of our VLFFD. The fine-grained prompt and the mixed forgery image are first generated via Prompt Forgery Image Generator (PFIG). Then the image encoder and text encoder are trained with the Coarse-and-Fine Co-training framework (C2F) inside the black dotted frame. The top half of the C2F is Coarse-grained Multimodal learning, while the bottom represents Fine-grained Multimodal learning.
        (Best viewed in color.)
       }
    \label{fig:main}
\end{figure*}
\subsection{Visual-Language Multimodal Leaning}
Vision and language are the two important signals for human perception, the visual-language multimodal learning thus has drawn lots of attention in the AI community.
Some works are devoted to using language information as a supervisory signal to guide vision tasks. Such as Visual Grounding~\cite{karpathy2014deep,kamath2021mdetr,zhu2022seqtr},
Vision-and-Language Navigation\cite{anderson2018vision,fried2018speaker},  
and Image Generation from Text~\cite{xu2018attngan,ramesh2021zero}. The success of these tasks demonstrate that language can help the vision models learn more fine-grained representations. Recently, another important multimodal learning paradigm \textit{i.e.} visual-language pretraining, has achieved great success. Specifically, CLIP~\cite{radford2021learning} first uses multimodal contrastive learning to train text and image encoders with 
4 million paired visual-language web data. Subsequently, many works have been put forward to fine-tune the VL pretrain model to adapt downstream tasks such as action recognition
\cite{wang2021actionclip}, and ReId\cite{yan2022clip}.

\section{Method}
\label{sec:method}
\begin{figure*}[!t]
    \begin{center}
    
       \includegraphics[width=1.0\linewidth]{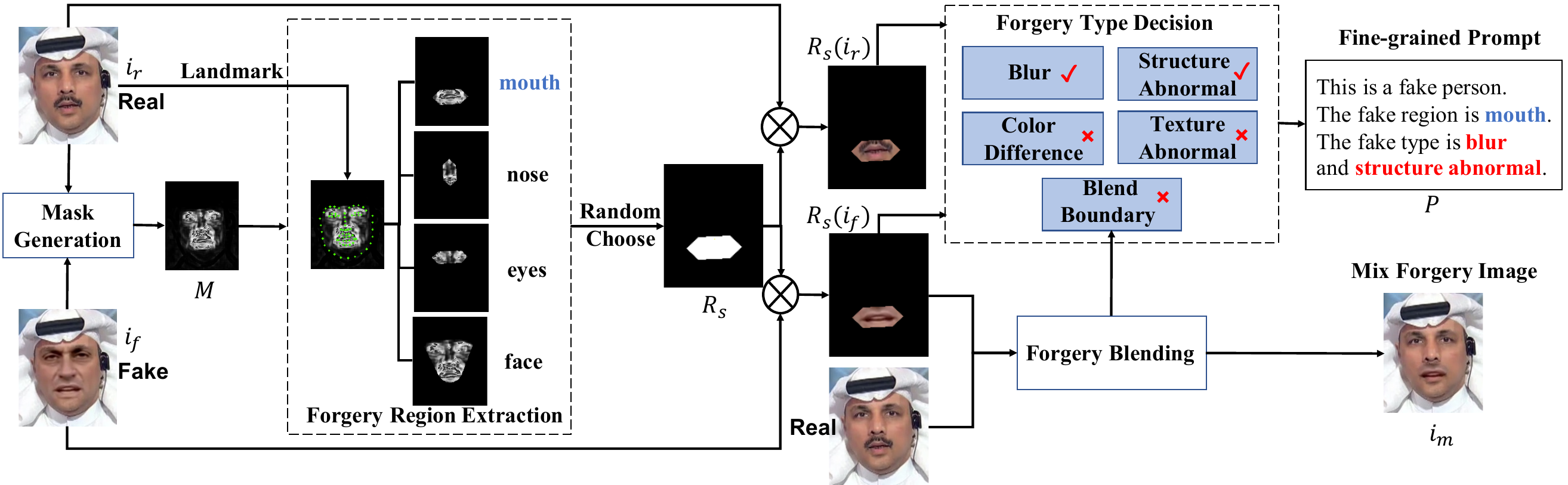}
    \end{center}
       \caption{ Overall framework of the Prompt Forgery Image Generator (PFIG). The paired forgery and real image are first fed into the Mask Generation module to generate forgery mask $M$. Then the Forgery Region Extraction module extracts the selected region $R_s$. Subsequently, the Forgery Type Decision module and Forgery Blending module decide the fine-grained forgery types of $R_s$ and generate the mixed forgery image, respectively.
       Finally, the fine-grained prompt is generated by the forgery region and types with the template. (Best viewed in color.)
       }
    \label{fig:pfig}
\end{figure*}
\begin{figure}[!t]
    \begin{center}
    
       \includegraphics[width=1\linewidth]{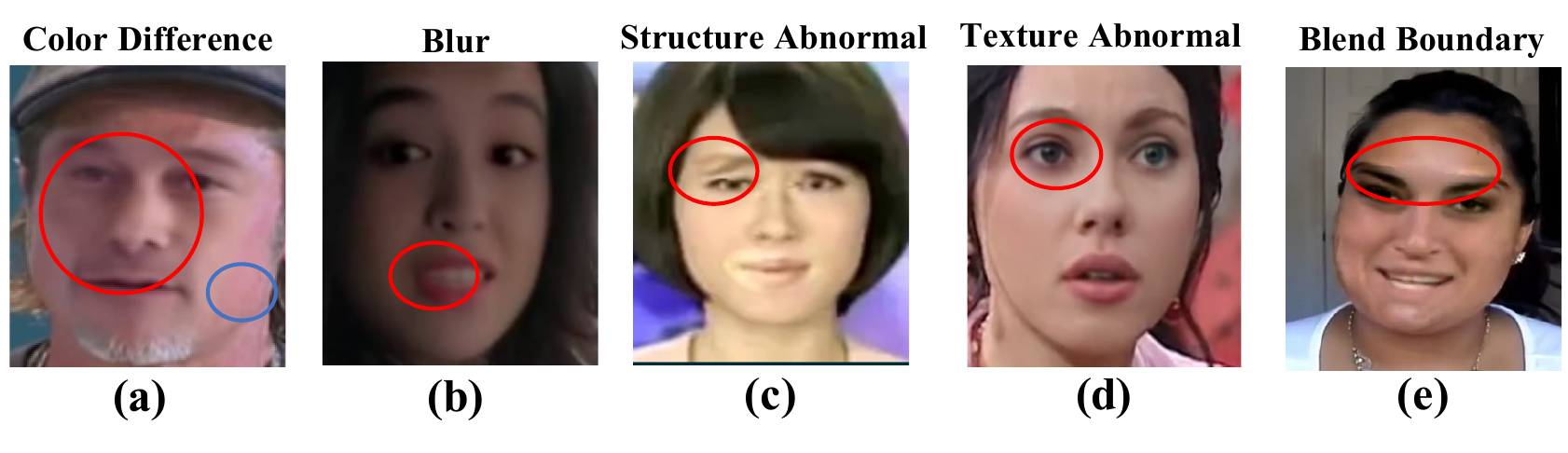}
    \end{center}
       \caption{Five typical types of forgery faces. (a) Color Difference. (b) Blur. (c) Structure Abnormal.
       (d) Texture Abnormal. (e) Blend Boundary.
       The red circle highlights the region of each forgery type. (Best viewed in color.)
       }
    \label{fig:type}
\end{figure}
In this section, we introduce our Visual-Linguistic Face Forgery Detection (VLFFD) framework, which 
explores the coarse-grained and
fine-grained natural language supervision with a pretrained multimodal framework to strengthen the generalization and interpretability. As shown in Fig \ref{fig:main}, VLFFD contains two unimodal encoders $E_i$ and $E_t$ for images and text prompts via dual stream framework. The $E_i$ extract image feature for the visual modality could be any architecture. The text encoder $E_t$ is used to extract language features of the input prompt. The output dimensions of $E_i$ and $E_t$ should be consistent in order to calculate the similarities.
The beginning of the VLFFD is
to generate the mixed forgery images with their corresponding fine-grained prompts using forgery dataset of paired real and fake faces, named Prompt Forgery Image Generator (PFIG). 
Subsequently, the Coarse-and-Fine Co-training Framework (C2F) is used to jointly train the coarse and fine data via multimodal contrastive learning, enabling encoders to potentially perform the coarse-to-fine learning process.

\subsection{Prompt Forgery Image Generator}
The existing face forgery dataset contains category-level labels (real or fake) while lacking sentence-level annotations. Collecting manual annotation not only
requires huge labor-intensive but also faces the problem of mislabeling.
Thus, in this work, we first propose \textit{Prompt Forgery Image Generator (PFIG)} to produce mixed forgery images with fine-grained annotation using an existing paired face forgery dataset. 
The PFIG disentangles the forgery artifacts from the existing dataset into \textbf{regions} and \textbf{types}, which has been argued that is two key elements of the forgery traces~\cite{chen2022self,shiohara2022detecting}. 
As shown in Fig~\ref{fig:pfig}, 
given a real image $i_r \in R^{ 3\times H \times W}$ with its corresponding forgery image $i_f \in R^{ 3\times H \times W}$, the PFIG include following steps:
    
    \textbf{Mask Generation}. To locate the forgery region, similar to~\cite{chen2021local}, we first construct manipulated mask $M$ by computing the absolute pixel-wise difference in the RGB channels, and normalized it to the range of $[0,1]$:
    \begin{equation}
        M = |i_r-i_f| / 255.
        \label{equation:1}
    \end{equation}
    
    \textbf{Forgery Region Extraction}. 
    This step aims to choose one forgery region that contains the $i_f$.
    Specially, we divide the facial image into four regions based on landmarks, including mouth, nose, eyes, and face. 
    Then, we calculate the average value of $M$ within each region area, and empirically determine the threshold $\theta$ to obtain the forgery regions list $L_f$. That is:
    \begin{equation}
        \frac{1}{|R_t|}\sum_{j\in R_t}M(j) > \theta, R_t \rightarrow L_f,
        \label{equation:2}
    \end{equation}
    where $R_t$ is one of the regional areas of the predefined four regions, $|R_t|$ represented the sum of pixels within the region $t$. If the value is greater than the $\theta$, the corresponding region is appended into regions list $L_f$.
    After looping the Eq.~\ref{equation:2} with four regions, we randomly select one region $R_s$ from $L_f$ for the next step. The $R_s(i_r)$ and $R_s(i_f)$ represent the forgery region of the real and fake pixels, respectively.
    
    \textbf{Forgery Type Decision}. The goal of this step is to determine the type of forgery via a specially designed criterion.
    According to the previous work and our observation, we categorize the existing forgery types as color difference, blur, structure abnormal, texture abnormal, and blend boundary. We detail each forgery type and corresponding evaluation standard as follows:
    1) \textbf{Color Difference}. 
    This phenomenon occurs in the face swap when the color of the source and target face has a drastic difference, as shown in Fig~\ref{fig:type}(a). Inspired by the color transfer~\cite{reinhard2001color},
    we leverage the distance of the average channel-wise mean and variance of the real and fake region in the $Lab$ color space to determine whether there exists a color difference. 2) \textbf{Blur}. To quantify the local blurring of forgery faces as shown in Fig~\ref{fig:type}(b), we exploit the Laplacian operator to reflect the sharpness of image edges. Specifically, we compute the variance of the real and fake images of the selected region after the Laplacian operator to determine whether this part is blurred.
    3) \textbf{Structure Abnormal}. We observed that some organs of fake faces will be obviously deformed, such as the left eye in Fig~\ref{fig:type}(c). To metric such a phenomenon, we use the Structural Similarity (SSIM) index difference between real and fake images of the selected region $R_s$ to decide whether the chosen region has a structure abnormal or not. 
    4) \textbf{Texture Abnormal}. It has been proved that the generator cannot generate as strong texture contrast as real data~\cite{liu2020global}, leading to texture difference as shown in Fig~\ref{fig:type}(d). Inspired by Gram-Net~\cite{liu2020global}, we leverage the contrast of Gray-Level Co-occurrence Matrix (GLCM)~\cite{haralick1973textural}, formed as $C_d$, to reflect the clarity of texture.
    We define a forgery region as texture abnormal when the $C_d$ of the real is larger than the fake one beyond the threshold.
    5) \textbf{Blend Boundary}. Existing face manipulation methods
    conduct blending operation to transfer an altered face into an existing background, which leaves intrinsic cues across the blending boundaries~\cite{li2020face}, such as the red circle of Fig~\ref{fig:type}(e). However, we find it difficult to directly detect the blend boundaries, especially if the forged image is of high quality. We thus enhance this trace by adjusting the blend methods and ratio in the next step.

    \textbf{Forgery Blending}. 
    To increase the variety of mixed forgery image $i_m$, we exploit Poisson blending or alpha blending according to a certain probability, formulated as follows:
    \begin{equation}
        \label{eq6}
        \left\{
        \begin{aligned}
            i_m =  & \alpha * R_s(i_f) + (1-\alpha)*i_r, &p<\theta_b\\
            i_m =  & Possion(R_s(i_f),i_r),& p>\theta_b,
        \end{aligned}
        \right.
    \end{equation}
    where $\alpha$ is the ratio used to control the degree of the blending, $Possion$ represents the standard Possion blending method, $p$ is a random variable sampled from a uniform distribution, $\theta_p$ is a threshold control the probability of two blending mechanisms. Compared with Poisson blending which leverages gradient regularization to smooth the boundary, alpha blending will bring obvious blending cues. Thus, we define the forgery type as Blend Boundary when choosing the alpha blending method.

Appendix provide pseudocodes for each forgery type criterion in detail. 
Finally, to combine the fine-grained annotation and adjust the up-stream pre-trained model, we generate sentence-level
fine-grained prompt $P$ for each mixed forgery image with template \textit{"this is a fake person, the forgery region is \_\_ , the forgery type is \_\_"}. The former blank fills with the selected forgery region $R_s$, the latter blank fills with the corresponding forgery types. Different from regular multimodal tasks that require raw descriptions of the image, recent works demonstrate that CLIP can achieve SOTA performance by simply filling templated text.
Following this paradigm, we use a specially designed cloze form as our prompt to fine-tune CLIP, which can better perceive the face regions and forgery types.

\subsection{Coarse-and-Fine Co-training Framework}
In this section, we introduce our learning framework adopted by the multimodal co-training mechanism. Specifically, the original data with coarse-grained prompts are trained by Coarse-grained Multimodal Learning (CML) to let $E_i$ and $E_t$ obtain the basic discrimination.
Meanwhile, Fine-grained Multimodal Learning (FML) further guides and refine the encoders to learn more generalized cues by learning from rich supervision semantic information. 

\noindent\textbf{Coarse-grained Multimodal Learning.}
To adapt the multimodal learning framework, we first convert the digitized binary labels (0 for real and 1 for forgery) into prompts. After trying several templates,
we choose \textit{"this is a fake person"} for the forgery image and \textit{"this is a real person"} for the real one as the coarse-grained language prompts. 
Then, the real and forgery prompts are fed into the text encoder $E_t$ and concentrated together to form a coarse-grained language feature, represents as $l_c \in R^{2 \times D}$. Given the input data $x \in R^{B\times 3 \times H \times W}$ (B is the batch size)
with label $y \in \{0,1\}$, the visual feature $v_c \in R^{B\times D} $ (D is the feature dimension) is obtained by the image encoder $E_i$. We use normalized cosine similarity as the metric of two modalities features denoted as $s(u,v) = \frac{u.v^T}{||u|||v||}$.
The coarse-grained loss $L_c$ based on cross-entropy loss can be defined as:
\begin{equation}
    L_{c} = -\frac{1}{B}\sum y log(s(v_c,l_c)).
\end{equation}

\noindent\textbf{Fine-grained Multimodal Learning.}  
To efficiently exploit the sentence-level knowledge, inspired by the visual-language pre-trained methods~\cite{radford2021learning}, we use the multimodal contrastive learning framework, which pulls the paired image and text representations close to each other.
Specifically, a batch of mixed forgery images $x_f \in R^{N \times 3 \times H \times W}$ ($N$ is the batch size) with prompt $p_f \in P$, are fed into the image encoder $E_i$ and text encoder $E_t$ to extract visual and language features $v_f \in R^{B \times D}$ and $l_f\in R^{B \times D}$, respectively, where $D$ is the dimension of the feature. Then, we compute the symmetric normalized cosine similarities between two modalities 
$s(v_f,l_f)$ and $s(l_f,v_f) \in R^{B\times B}$. The diagonal elements of the two symmetric matrices represent the similarity of paired features, while others are mismatched. Thus, we define the symmetric cross-entropy loss as the fine-grained loss $L_{f}$, which maximizes the diagonal of the two similarity matrices as follows:
\begin{equation}
    L_{f} = -\frac{1}{2N}(\sum log(s(v_f,l_f) \odot I) + \sum log(s(l_f,v_f) \odot I)),
\end{equation}
where $I \in R^{N \times N}$ represents the identity matrix. Since the beginning of the fine-grained prompt is the forgery coarse-fined prompt, the $L_f$ can not only 
help encoders extract more discriminative representations but also encourage the image encoder to extract more refined features that promote generalization.

\noindent\textbf{Overall Loss Function.} Considering both the
coarse-grained loss and fine-grained loss, the overall loss for our proposed framework is:
\begin{equation}
    L = L_c + \phi L_f,
\end{equation}
where $\phi$ is the hyperparameter used to scale the fine-grained loss. 

\section{Experiment}
\label{sec:experiment}
\subsection{Experimental Setting}

\begin{table*}[!h]
    \renewcommand\arraystretch{1.1}
    \centering
    \resizebox{0.95\textwidth}{!}{
    \begin{tabular}{c|cc|cccccccc}
    \hline
    \multirow{2}*{Method} & \multicolumn{2}{c|}{\textit{FF++}}&\multicolumn{2}{c}{DFD} & \multicolumn{2}{c}{DFDC-P} & \multicolumn{2}{c}{Wild Deepfake} &\multicolumn{2}{c}{Celeb-DF}\\
    \cmidrule{2-3}
    \cmidrule{3-11}

    & AUC& EER &AUC& EER& AUC& EER& AUC& EER& AUC& EER\\
    \hline
    Xception~\cite{chollet2017xception}& 99.09&3.77 &87.86& 21.04& 69.80& 35.41&66.17 &40.14 & 65.27& 38.77\\
    EN-b4~\cite{tan2019efficientnet}   &99.22 &3.36& 87.37      & 21.99      & 70.12       & 34.54      & 61.04           & 45.34           & 68.52         & 35.61        \\
    Face X-ray~\cite{tan2019efficientnet}& 87.40&-& 85.60&-& 70.00&      -      &       -          &         -        & 74.20          &       -       \\
    
    F3-Net~\cite{qian2020thinking}          & 98.10& 3.58 & 86.10     & 26.17       &     72.88       &      33.38      & 67.71          &      40.17           & 71.21        &     34.03         \\
    MAT~\cite{zhao2021multi} & 99.27&3.35& 87.58      & 21.73      & 67.34       & 38.31      & 70.15           & 36.53           & 70.65         & 35.83        \\
    GFF~\cite{luo2021generalizing}             & 98.36& 3.85 & 85.51      & 25.64      & 71.58       & 34.77      & 66.51           & 41.52            & 75.31         & 32.48        \\
    LTW~\cite{sun2021domain}            & 99.17& 3.32  & 88.56      & 20.57      & 74.58       & 33.81      & 67.12           & 39.22           & 77.14         & 29.34         \\
    
    LRL~\cite{chen2021local}    & \textbf{99.46}&\textbf{3.01}& 89.24      & 20.32      & 76.53       & 32.41      & 68.76           & 37.50              & 78.26         & 29.67        \\
    
    DCL~\cite{sun2021dual}              & 99.30&3.26& 91.66      & 16.63      & 76.71       & 31.97      & 71.14         & 36.17           & 82.30       & 26.53   \\
    PCL+I2G~\cite{zhao2021learning}& 99.11&-& -      & -      & -       & -      & -         & -           & 81.80       & -   \\
    SBI~\cite{shiohara2022detecting}    & 88.33&20.47&88.13&17.25&76.53&30.22&68.22&38.11&80.76&26.97 \\
    UIA-ViT~\cite{zhuang2022uia} &-&-&94.68&-&75.80&-&-&-&82.41&- \\

    \hline

    Ours              & 99.23&3.12& \textbf{94.79}      & \textbf{15.31}      & \textbf{84.74}       & \textbf{23.43}      & \textbf{83.55}           & \textbf{24.40}           & \textbf{84.80}          & \textbf{22.73}       \\
    \hline

    \end{tabular}
    }
    \caption{Frame-level cross-database evaluation from FF++(HQ) to DFD, DFDC, Wild Deepfake and Celeb-DF in terms of AUC and EER. The FF++ belongs to the intra-domain results while others represent to the unseen-domain.
    }
    \label{table:1}
    \end{table*}

\noindent\textbf{Dataset.}
To evaluate the generalization of our proposed VLFFD, we 
conduct our experiments on several challenging datasets:
1) FaceForensics++~\cite{rossler2019faceforensics++}: a widely-used forgery dataset contains 1000 videos with four different manipulated approaches, including two deep learning based \textit{DeepFakes} and \textit{NeuralTextures} and two graphics-based methods \textit{Face2Face} and \textit{FaceSwap}. This dataset provides  pairwise real and forgery data, enabling us to generate mixed forgery images with PFIG. 2) DFDC-P~\cite{dolhansky2020deepfake} dataset is a challenging dataset with $1133$ real videos and $4080$ fake videos, containing various manipulated methods and backgrounds.
3) DFD is a forgery dataset containing $363$ real videos and $3068$ fake videos, which is mostly generated by the Deepfake method. 4) Celeb-DF~\cite{li2019celeb}
is another high-quality Deepfake dataset that contains various scenarios. 5) Wild-Deepfake~\cite{zi2020wilddeepfake} is a forgery face dataset
obtained from the internet, leading to a diversified distribution of scenarios. We use DSFD~\cite{li2019dsfd} to extract faces from each video.


\noindent\textbf{PFIG details.}
We use the open-source dlib algorithm as the face landmark detector. For the forgery type decision, the threshold of mean and variance is $1.0$ and $0.5$. 
For the blur, the threshold is set to $100$. If the difference of SSIM is larger than $0.97$, we determine the forgery part is structure abnormal. The texture abnormal threshold is set to $0.7$. The blending ratio $\alpha$ is set to $0.9$.

\noindent\textbf{Training details.}
To leverage the visual-language correspondence information,
we use CLIP\cite{radford2021learning} as the pretrain model of $E_i$ and $E_t$. The feature dimension $D$ is $768$. 
We resize the input into $224\times 224$, and the ViT-L is used as the image encoder. We use Adam optimizer to train the framework and the learning rate is set to $1e-6$. The batch size of Coarse-grained Multimodal Leaning $B$ is set to $32$ and the batch size of Fine-grained Multimodal Leaning $N$ is set to $24$. The hyperparameters $\phi$ is set to $0.1$. The total training epoch is 40 and the overall framework is implemented in PyTorch on one NVIDIA A-100 GPU.

\noindent\textbf{Testing details.}
During testing, Following CLIP~\cite{radford2021learning}, all text features corresponding to coarse-grained and fine-grained prompts can be extracted in advance, and the image feature of the test face is matched with the most similar text feature by cosine similarity to obtain the binary label as well as the text description. The test period can be divided into coarse-grained classification and fine-grained matching. The former aims to identify the real or fake of the input face, while the latter output the sentence-level description in terms of forgery region and types. All the quantization results use the coarse-grained classification.
\subsection{Experimental Results}
\noindent\textbf{Cross-dataset evaluation.} To demonstrate the generalization of our VLFFD, we first evaluate performance on the unseen datasets against the recent SOTA methods.
Specifically, we train our model on the high-quality version of FF++ and test on the other datasets, which have large domain gaps with FF++ in terms of forgery types, human ids, video backgrounds, and image quality \textit{e.t.c}. 
The quantitative frame-level results are shown in Tab~\ref{table:1}. We can observe that our method can obtain significant improvement compared with the other methods in terms of AUC and EER.
Concretely, our methods outperform about 8\% on DFDC-P and 12\% on Wild-Deepfake compared with framework engineering based methods DCL.
Compared with forgery simulation based SBI and PCL+I2G, VLFFD can achieve better performance on Celeb-DF.
Furthermore, our method outperforms the recent transformer-based methods UIA-ViT by 9\% and 2\% on DFDC-P and Celeb-DF, respectively.
The results demonstrate that fine-grained language information and a powerful visual-language pre-training model can greatly improve the generalization ability.

\noindent\textbf{Cross-manipulation evaluation.}
The aforementioned experiments show the effectiveness of VLFFD under large domain gap situations. To further demonstrate the generalization among different manipulated methods, we train a model on one method within the high-quality of FF++ dataset and test on the four methods. We compare with the recent four methods including the Multi-attentional, GFF and DCL, results for these methods are from~\cite{sun2021dual}.
As shown in Tab~\ref{table:2}, our proposed method achieve the SOTA performance in most situations. In particular, in several cases, the VLFFD outperforms the compared methods significantly. For example, when training on the FaceSwap and testing on the DeepFakes, the performance improved by over 18\% in terms of AUC against DCL. The performance gain because the fine-grained supervisory information guides the model to capture more refined forgery patterns, which are shared across the different forgery types, such as the texture abnormal face in Deepfakes.

\begin{table}[!t]
    \centering
    \renewcommand\arraystretch{1.1}
    \resizebox{1,0\columnwidth}{!}{
    \begin{tabular}{c|c|cccc}
    \hline
    Train   & Method       & DF    & F2F   & FS    & NT    \\
    \hline
    \multirow{4}{*}{DF}  
    & MAT          & \textit{99.92} & 75.23 & 40.61 & 71.08 \\
    & GFF          & \textit{99.87}	&76.89	&47.21&	72.88 \\
    & DCL    & \textit{\textbf{99.98}} & 77.13 & 61.01 & 75.01 \\
    & Ours    & \textit{99.97} & \textbf{87.46} & \textbf{74.40} & \textbf{76.79} \\
    \hline
    \multirow{3}{*}{F2F} 
    & MAT          & 86.15&	\textit{99.13}&	60.14&	64.59 \\
    & GFF          & 89.23	&\textit{99.10}&	61.30&	64.77 \\
    & DCL    & 91.91  & \textit{99.21} & 59.58 & 66.67 \\
    & Ours    & \textbf{94.90} & \textit{\textbf{99.30}} & \textbf{65.19} & \textbf{66.69} \\
    \hline
    \multirow{3}{*}{FS}  
    & MAT          & 64.13	&66.39&\textit{99.67}	&	50.10 \\
    & GFF          &70.21&	68.72 &	\textit{99.85}&	49.91 \\
    & DCL    & 74.80  &   69.75&\textit{99.90}  &52.60  \\
    & Ours    & \textbf{92.98} & \textbf{79.69} & \textit{\textbf{99.57}} & \textbf{53.53} \\
    \hline
    \multirow{3}{*}{NT}  
    & MAT          &  87.23	&48.22&	75.33&	\textit{98.66}  \\
    & GFF          & 88.49	&49.81&	74.31&	\textit{98.77} \\
    & DCL    & 91.23 & 52.13 &79.31 & \textit{98.97}\\
    & Ours    & \textbf{92.74} & \textbf{62.53} & \textbf{85.62} & \textit{\textbf{98.99}} \\
    \hline
                         
    \end{tabular}
    }
    \caption{Cross-manipulation evaluation in terms of AUC. Diagonal results indicate the intra-domain performance.
    }
    \label{table:2}

    \end{table}


\subsection{Ablation Study}
In this section, we perform an extensive ablation study to explore the impact of each component of the VLFFD. 

\noindent\textbf{Impact of different components.}
To analyze the impact of the Multimodal learning and PFIG module, we adapt the image-encoder of CLIP to the binary classification task and remove the text encoder, which has the same backbone and pretrain weight as our multimodal version. 
The results are shown in Tab~\ref{table:ab1}. The PFIG represents whether use the generated mixed forgery images to train the model.
We can observe a consistent improvement in performance when using the multimodal learning framework, with or without PFIG, which demonstrates the advantage of language information with the visual-linguistic pre-training model.
Furthermore, the introduction of PFIG leading a 3\% increase in AUC, which demonstrates the effectiveness of fine-grained data generated by PFIG.

\begin{table}[!t]
    \centering
    \renewcommand\arraystretch{1.2}
    \resizebox{\columnwidth}{!}{
    \begin{tabular}{c|c|cc|cc}
        \hline
        \multirow{2}*{Multimodal}&\multirow{2}*{PFIG} & \multicolumn{2}{c|}{Celeb-DF}&
        \multicolumn{2}{c}{DFDC-P}  \\

        \cline{3-6}
        && AUC& EER &AUC& EER \\
        \hline
        $\times$&$\times$&77.62&29.34&76.31&31.06\\
        $\times$&$\checkmark$&77.71&29.31&77.30&29.83\\
        $\checkmark$&$\times$&80.79&25.16&81.79&26.13\\
        $\checkmark$&$\checkmark$&\textbf{84.80}&\textbf{22.73}&\textbf{84.74}&\textbf{23.43}\\
        \hline
    \end{tabular}
    }
    \vspace{-2pt}
    \caption{Ablation study on the impact of different components in terms of AUC and EER. The Multimodal represents whether convert the digital label into prompt and use the text-encoder. 
    }
    \label{table:ab1}
\end{table}

\begin{table}[!t]
    \centering
    \renewcommand\arraystretch{1.2}
    \resizebox{\columnwidth}{!}{
    \begin{tabular}{c|c|c|cc|cc}
        \hline
        \multirow{2}*{Backbone}&\multirow{2}*{CLIP}&\multirow{2}*{VLFFD} & \multicolumn{2}{c|}{WDF}&
        \multicolumn{2}{c}{DFDC-P}  \\

        \cline{4-7}

        &&& AUC& EER &AUC& EER \\
        \hline
        \multirow{2}*{Xception}&$\times$&$\times$&66.17&40.14&69.80&35.41\\
        &$\times$&$\checkmark$&\textbf{70.22}&\textbf{37.37}&\textbf{76.11}&\textbf{30.65}\\
        \cline{1-7}

        \multirow{2}*{EN-B4}&$\times$&$\times$&61.04&45.34&70.12&34.54\\
        &$\times$&$\checkmark$&\textbf{71.35}&\textbf{36.22}&\textbf{78.32}&\textbf{29.03}\\
        \hline
        \multirow{3}*{ViT-B}&$\times$&$\times$&65.39&38.11&68.03&36.06\\
        &$\times$&$\checkmark$&74.64&30.94&75.31&30.69\\
        &$\checkmark$&$\checkmark$&\textbf{78.80}&\textbf{27.81}&\textbf{78.00}&\textbf{29.60}\\
        \cline{1-7}
        \multirow{3}*{ViT-L}&$\times$&$\times$&69.92&36.20&71.33&33.76\\
        &$\times$&$\checkmark$&76.49&30.49&79.53&28.66\\
        &$\checkmark$&$\checkmark$&\textbf{84.80}&\textbf{22.73}&\textbf{84.74}&\textbf{23.43}\\
        \hline
    \end{tabular}
    }
    \vspace{-2pt}
    \caption{Ablation of the pretrain weight and backbones. CLIP indicates whether the image encoder uses CLIP based pretrain weight.}
    \label{table:ab2}
\end{table}

\begin{figure*}[!t]
    \begin{center}
    
       \includegraphics[width=1\linewidth]{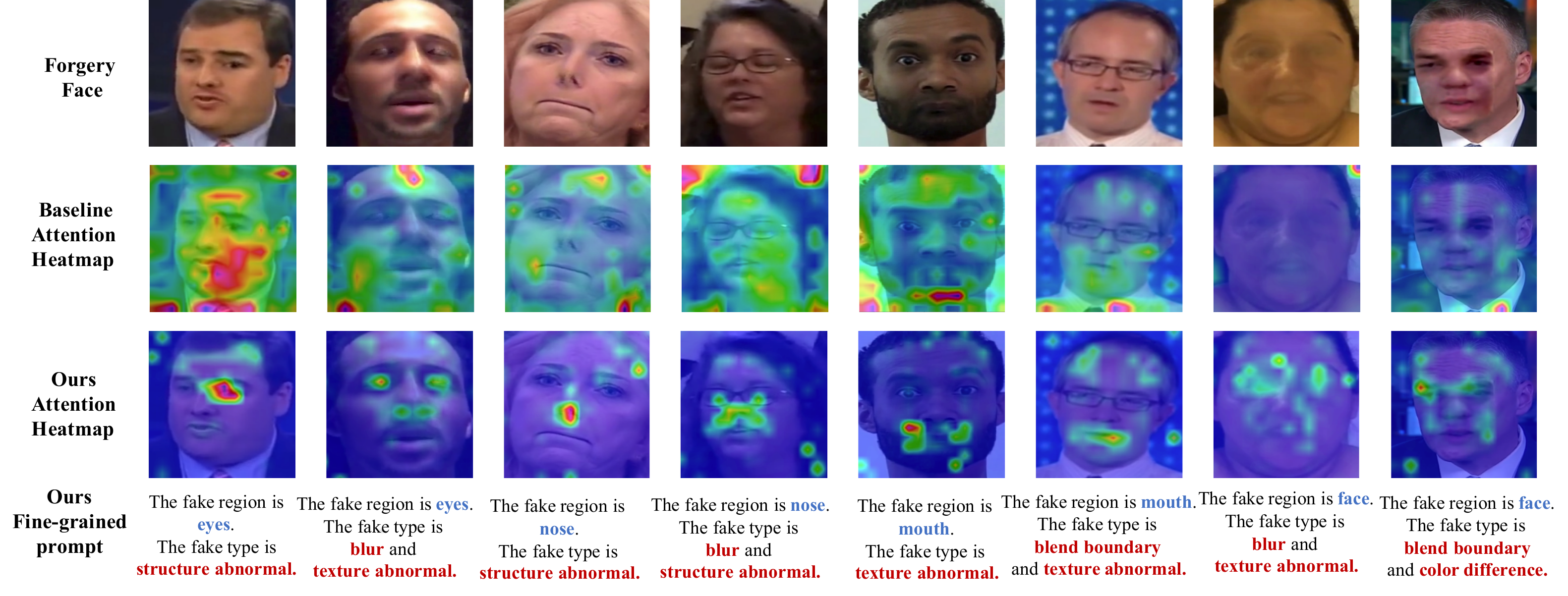}
    \end{center}
       \caption{ Attention heatmap visualization of the baseline and our model. The first row represents the original fake images that did not appear in the training set.
       The last row represents the Top-1 matching prompts of our methods. More visualization results are provided in the supplementary material.
       (Best viewed in color.)
       }
    \label{fig:vis1}
\end{figure*}

\noindent\textbf{Impact of pretrain weight and backbone.}
To analyze the impact of pretrain weight and backbone, we explore pretrain weight from ImageNet as well as different backbones, including CNN-based backbones Xception and EN-B4 and Transformer-based backbones ViT-B and ViT-L.
To adapt the CNN-based backbone into our framework, we use MLP to align the feature dimension with the CLIP pretrain text encoder.
The quantitative results are shown in Tab~\ref{table:ab2}, we can observe that 1) the performance is consistently improved after adopting our VLFFD into different backbones, which demonstrates that our method can boost generalization even without pretrain knowledge.
2) The CLIP-based image encoder outperforms ImageNet-based by 4\% and 5\% on Celeb-DF and DFDC-P datasets, respectively. That may prove that the semantic knowledge hidden in the VL pretrain model can better unleash the potential of our approach.

       



\begin{table}[!t]
   \centering
   \renewcommand\arraystretch{1.2}
   \resizebox{1.0\columnwidth}{!}{
   \begin{tabular}{c|c|c|c|c|c}
       \hline
       
        \multirow{2}*{Siginal}&\multirow{2}*{Method} &\multicolumn{2}{c|}{Celeb-DF}&\multicolumn{2}{c}{DFDC-P}\\
       \cline{3-6}

       && AUC &EER& AUC &EER\\
       \hline
       \multirow{1}*{Mask}&Decoder&77.70&29.56&78.51&29.59\\
       \hline
       \multirow{3}*{Digital}&Region(MC=4)&79.73&29.24&78.59&29.07\\
       &Type(MC=31)&78.45&29.95&78.17&29.92\\
       &Region+Type(MC=101)&77.18&30.47&77.54&31.03\\
       \hline
       
       \multirow{3}*{Text}&Region&\textbf{82.26}&\textbf{25.58}&\textbf{82.37}&\textbf{24.35}\\
       &Type&\textbf{81.04}&\textbf{26.73}&\textbf{82.09}&\textbf{24.91}\\
       &Region+Type&\textbf{84.80}&\textbf{22.73}&\textbf{84.74}&\textbf{23.43}\\

       \hline
   \end{tabular}
   }
   \caption{Ablation study on different supervisory signals. MC is represented as multiple classifiers.}
   \label{table:ablation}
\end{table}
\noindent\textbf{Impact of language information.}
To quantify the importance of language information, we compare our text-based supervisory signals with masks and digital labels. We use a four-layer upsampling convolution as a decoder after the last feature of the image-encoder to regress on the forgery mask and compute the mean squared error loss with the mask obtained from Eq.~\ref{equation:1}. For the digital labels, we replace the text-encoder with multiple classifiers (MC) that have three levels of granularity (Region only, Type only, and both Region and Type) for the generated mixed forgery samples. As shown in Tab~\ref{table:ablation}, our text-based method surpasses both mask and multiple classifiers for all levels of granularity. The mask-supervised method exhibits weak generalization, suggesting that mask supervision alone can cause overfitting to the training data. Moreover, unlike the text-encoder that improves steadily with more fine-grained supervision, the multiple classifiers are affected by the number of classes and class imbalance problem, which further confirms the scalability and versatility of language over multi-classification.

\subsection{Visualizations}

    

\noindent\textbf{Visualizations of classification decision.}
To understand the decision-making mechanism of our method, we visualize the attention heatmap of the baseline method (binary classification with CLIP pretrained image-encoder) and our VLFFD for the unseen test forgery image, respectively. Furthermore, we also show the Top-1 similarity
score of fine-grained prompt as the linguistic interpretation of our model. As shown in Fig~\ref{fig:vis1}, we can observe that the attention is more clearly directed to the fine-grained
forgery regions compared with the baseline. In addition, the highlighted attention area basically corresponds to the region in the matched semantic sentence. For example, the attention of the baseline method ignores the abnormal region of the right eye of the first face. After training with the VLFFD framework, the network clearly focuses on this region due to the supervision of fine-grained semantic information. The prompt can further provide a detailed classification basis for regions and types.
This visualization demonstrates that our VLFFD can effectively capture more refined and accurate forgery features and improve the interpretability of the model. More visualizations are provided in the supplementary material.



\begin{figure*}[!t]
    \begin{center}
       \includegraphics[width=1\linewidth]
       {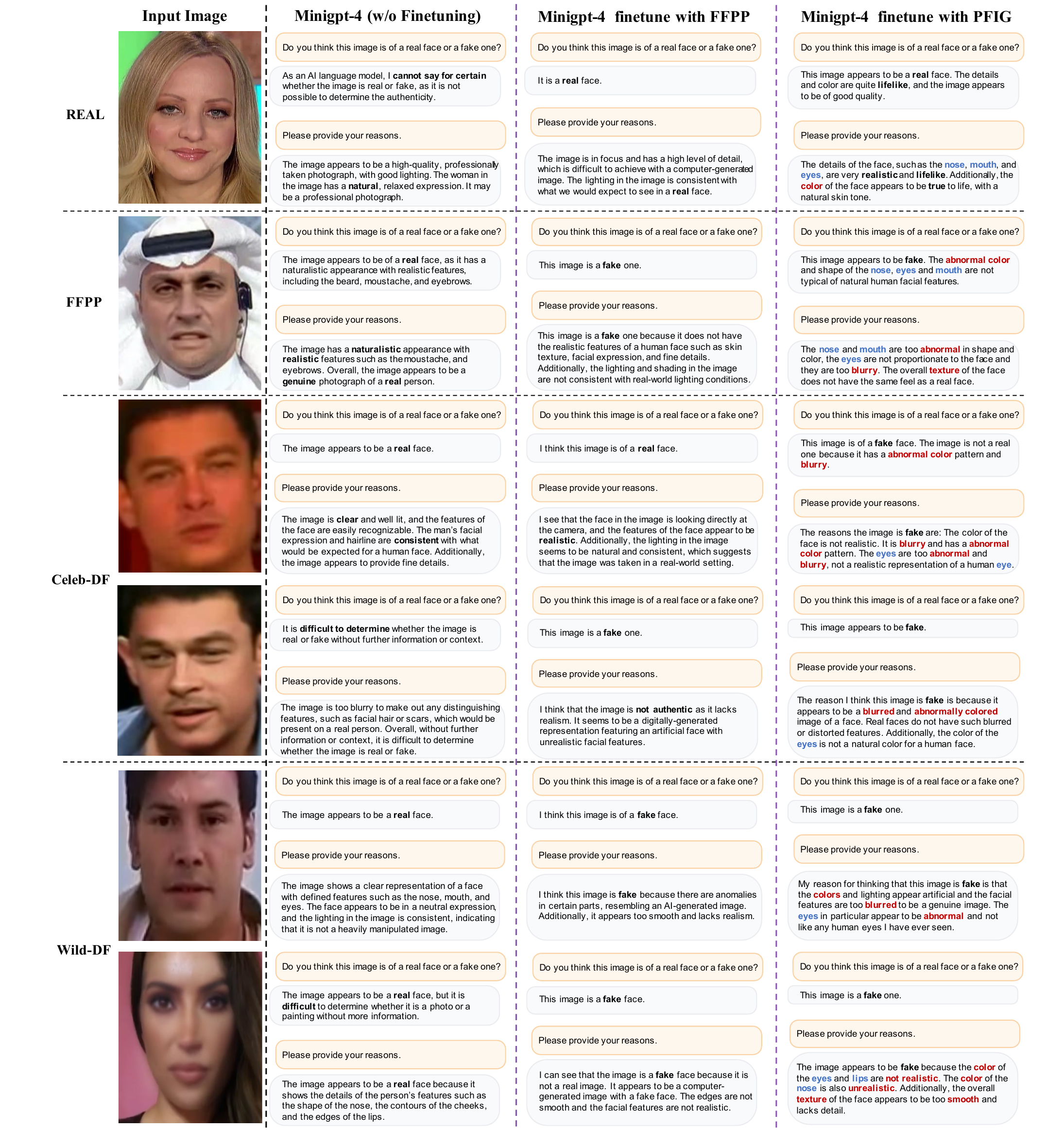}
    \end{center}
    
       \caption{Visualization of responses from different version of MiniGPT-4. Bolded font indicates the basis of the model's judgment, while red and blue colors signify words related to region and type, respectively. 
       }
    \label{fig:fakegpt}
\end{figure*}

\section{Integrating with Multimodal LLM}
\label{sec:D}
Recently, Multimodal Large Language Models (LLMs) have become a topic of interest in various research fields. Their ability to process and interpret complex data has proven crucial for tasks involving decision-making and understanding intricate datasets. Yet, the field of face forgery detection has not fully capitalized on this advancement, hindered by a lack of suitable image-text datasets to train such comprehensive models. Our work introduces the Prompt Forgery Image Generator (PFIG) method, which provides a strategy to generate fine-grained, annotated image-text pairs for forgery data. This dataset can be instrumental in training or fine-tuning multimodal LLMs, potentially enabling them to distinguish between authentic and forged images and to provide reasoning for their determinations.

To explore the potential of multimodal LLMs for face forgery detection, we fine-tuned the MiniGPT-4~\cite{zhu2023minigpt} model using the image-text pairs generated by our PFIG module. MiniGPT-4 is a compact yet powerful model designed to blend visual and linguistic data efficiently, which boasts capabilities like creating detailed visual descriptions but with a focus on computational efficiency and streamlined architecture. The adaptability of MiniGPT-4 to seamlessly blend and interpret this complex, multimodal data is a testament to its architectural ingenuity. It leverages a streamlined design that emphasizes computational efficiency, enabling it to generate precise visual descriptions and detect subtle inconsistencies in deepfake images without the extensive resource requirements typical of larger models.


Specifically, our approach entailed employing the PFIG module to create mixed forgery images accompanied by prompts. In addition, we generated fine-grained textual descriptions for the original FFPP dataset using the same PFIG module. These two components together formed the dataset used for fine-tuning. Due to training resource constraints, we adapted the MiniGPT-4 with the image encoder, Q-former module, and projection layer that interfaces with the LLM. 
We use the following questions as the prompt during the training period :(1). \textit{Do you think this image is of a real face or an altered fake one? Please provide your reasons.} (2). \textit{Do you believe this image shows an authentic human face, or is it a manipulated counterfeit? Please state your reasoning.} (3). \textit{Regarding this image, would you argue that it represents a genuine face or a tampered false one? Please explain your rationale.}
The learning rate is set to 3e-5, and the image size is resized into $224\times 224$.


We evaluated the models by posing the question: \textit{Do you think this image is of a real face or a fake one?}" followed by \textit{"Please provide your reasons.} for further explanation. As shown in Fig~\ref{fig:fakegpt},
we analyzed two variants of MiniGPT-4: (a) original MiniGPT-4 without fine-tuning and (b) MiniGPT-4 fine-tune with FFpp dataset with coarse label (``This is a fake person" for fake and ``This is a real person" for real). The initial model's vague responses highlighted its limited ability to differentiate forged faces. Furthermore, training with simple coarse labels from existing datasets did not guarantee accuracy or provide clear justifications. In contrast, the model refined with our PFIG data not only pinpointed authenticity but also delivered comprehensive explanations.
In summary, the LLM fine-tuned with our PFIG module shows improved judgment capability, supporting the potential positive impact of our method on face forgery detection LLM models. In the future, we plan to employ more extensive datasets and advanced multimodal LLMs for fine-tuning to create an even more effective forgery detection LLMs.

\section{Conclusion}
\label{sec:conclusion}
This paper focuses on improving the generalization and interpretability of the face forgery detection task via visual-linguistic manner. Traditional methods usually use binary numbers label as the supervisory signal which lacks fine-grained guidance and semantic information. In this paper, we tackle this issue by proposing a novel multimodal learning paradigm named VLFFD, which introduces the language modality as fine-grained semantic supervisory signals. Specifically, we first leverage PFIG module to automatically generates mixed forgery images with fine-grained prompts by analyzing the original forgery dataset. Then the C2F is designed to learn the multi-grained semantic information. Extensive experiments demonstrate the significant superiority of our method over state-of-the-art methods.

\section{Acknowledgments}
\label{sec:ack}
This work was supported by National Key R\&D Program of China (No.2022ZD0118202), the National Science Fund for Distinguished Young Scholars (No.62025603), the National Natural Science Foundation of China (No. U21B2037, No. U22B2051, No. 62176222, No. 62176223, No. 62176226, No. 62072386, No. 62072387, No. 62072389, No. 62002305 and No. 62272401), and the Natural Science Foundation of Fujian Province of China (No.2021J01002,  No.2022J06001).

{\small
\bibliographystyle{ieee_fullname}
\bibliography{main}
}
\clearpage
\appendix

\section*{\centering \Large Appendix \label{appendix}}

\section{Details of PFIG.}
\noindent\textbf{Color Difference.}
This phenomenon occurs in the face swap when the color of the source and target face has a drastic difference. Inspired by the color transfer~\cite{reinhard2001color},
we leverage the distance of the average channel-wise mean and variance of the real and fake regions in the $Lab$ color space to determine whether there exists a color difference. The $Lab$ color space minimizes correlation between channels, which helps reduce the impact of changes in a certain channel on the overall color.
The pseudocode is shown in Alg.~\ref{alg:cd}, $split$ represents dividing the channel of the image, $Lab$ denotes converting the RGB color space into $Lab$ space.

\begin{algorithm}[!h]
    
    \caption{Color Difference Decision}
    \begin{algorithmic}[1]
        \Require{Real image selected region $R_s(i_r)$, fake image selected region $R_s(i_f)$, mean threshold $\theta_c^m$, standard deviation threshold$\theta_c^s$}
        \State $R_s(i_r)^{'},R_s(i_f)^{'} = Lab(R_s(i_r)),Lab(R_s(i_f))$
        \State $L_r,a_r,b_r = split(R_s(i_r)^{'})$
        \State $L_f,a_f,b_f = split(R_s(i_f)^{'})$
        \State $L^m = ||mean(L_r) - mean(L_f)||_2$
        \State $a^m = ||mean(a_r) - mean(a_f)||_2$
        \State $b^m = ||mean(b_r) - mean(b_f)||_2$
        \State $L^s = ||std(L_r) - std(L_f)||_2$
        \State $a^s = ||std(a_r) - std(a_f)||_2$
        \State $b^s = ||std(b_r) - std(b_f)||_2$
        \State m = ($L^m$ + $a^m$ + $b^m$) / 3
        \State s = ($L^s$ + $a^s$ + $b^s$) / 3
        \If {m $> \theta_c^m$ and s $> \theta_c^s$}
            \State \textbf{Return} True
        \Else
            \State \textbf{Return} False
        \EndIf
    \end{algorithmic}
    \label{alg:cd}
    
\end{algorithm}

\noindent\textbf{Blur.}
There exists local blurring in forgery faces due to the instability of the generated model or blending operation. To quantify such phenomena, we make use of the Laplacian image, which can reflect the sharpness of image edges.
Specifically, as shown in Alg.~\ref{alg:blur}, we compute the variance of the real and fake images of the selected region after the Laplacian operator, and if the value of the real is larger than the fake one and their difference is greater than a certain threshold, we define this part as blurred. The $Laplacian(.)$ represents the Laplacian operator, $var(.)$ means calculating the variance of the input image.

\begin{algorithm}[!h]
    
    \caption{Blur Decision}
    \begin{algorithmic}[1]
        \Require{Real image selected region $R_s(i_r)$, fake image selected region $R_s(i_f)$, variance threshold $\theta_b^v$}
        \State $r\_var = var(Laplacian(R_s(i_r)))$
        \State $f\_var = var(Laplacian(R_s(i_f)))$

        \If {$r\_var > f\_var$ and ($r\_var - f\_var) > \theta_b^v$}
            \State \textbf{Return} True
        \Else
            \State \textbf{Return} False
        \EndIf
    \end{algorithmic}
    \label{alg:blur}
    
\end{algorithm}

\noindent\textbf{Structure Abnormal.}
We observed that compared with normal faces, some organs of fake faces will be obviously deformed. To metric such structure deformable, we use the Structural Similarity (SSIM) index difference between real and fake images of the selected region $R_s$ to decide whether the chosen region has a structure abnormal or not, which details in Alg.~\ref{alg:sab}.

\begin{algorithm}[!h]
    
    \caption{Structure Abnormal Decision}
    \begin{algorithmic}[1]
        \Require{Real image selected region $R_s(i_r)$, fake image selected region $R_s(i_f)$, ssim threshold $\theta_s$}
        \State $s = ssim(R_s(i_r),R_s(i_f))$
        \If {$s < \theta_s$}
            \State \textbf{Return} True
        \Else
            \State \textbf{Return} False
        \EndIf
    \end{algorithmic}
    \label{alg:sab}
    
\end{algorithm}

\noindent\textbf{Texture Abnormal.}
It has been proved that the generator typically correlates the values of nearby pixels and cannot generate as strong texture contrast as real data~\cite{liu2020global}, leading to texture differences in some forgery regions. Similar to the Gram-Net~\cite{liu2020global}, we leverage a texture analysis tool--the contrast of Gray-Level Co-occurrence Matrix (GLCM)~\cite{haralick1973textural}, formed as $C_d$.
Larger $C_d$ reflects stronger texture contrast, sharper and clearer visual effects. Inversely, a low value $C_d$ means the texture is blurred and unclear.
We define a forgery region as texture abnormal when the $C_d$ of the real is larger than the fake one beyond the threshold. The algorithm is shown in Alg.~\ref{alg:tab}, where $GLCM$ represents the average Gray-Level Co-occurrence Matrix of the input from right, down, left, and upper four orthogonal directions.

\begin{algorithm}[!h]
    
    \caption{Texture Abnormal Decision}
    \begin{algorithmic}[1]
        \Require{Real image selected region $R_s(i_r)$, fake image selected region $R_s(i_f)$, contrast threshold $\theta_t$}
        \Ensure{ $N=256 \times 256$}
        \State $P_r = GLCM(R_s(i_r))$
        \State $P_f = GLCM(R_s(i_f))$
        \State $C_d^r = \frac{1}{N}\sum_{i=0}^{255}\sum_{j=0}^{255}|i-j|^2P_r(i,j)$
        \State $C_d^f = \frac{1}{N}\sum_{i=0}^{255}\sum_{j=0}^{255}|i-j|^2P_f(i,j)$

        \If {$C_f^r > C_d^f$ and ($C_d^r - C_d^f) > \theta_t$}
            \State \textbf{Return} True
        \Else
            \State \textbf{Return} False
        \EndIf
    \end{algorithmic}
    \label{alg:tab}
    
\end{algorithm}

\section{Additional Experimental Results.}
\noindent\textbf{Multi-source manipulation evaluation.}
As mentioned in~\cite{sun2021domain}, the multisource cross-manipulation has great practical significance. Therefore, we also test performance when training on three forgery methods and test on the unknown forgery method. Furthermore, to demonstrate the robustness of our method against different image qualities, we conduct these experiments on both high-quality and low-quality datasets. The results are reported in Tab.~\ref{table:4}. Our method achieves the SOTA performance on all protocols and quality. Specifically, our method outperforms UIA-ViT by 4\% in the high-quality version of DF.
Furthermore, our method achieves significant improvement compared with recent DCL in low-quality settings,
which shows the robustness and generalization of our proposed framework.
\begin{table*}[!t]
    \centering
    \renewcommand\arraystretch{1.1}
    \resizebox{0.9\textwidth}{!}{
        \scalebox{1}{
    \begin{tabular}{c|cc|cc|cc|cc}
        \hline
        \multirow{2}*{Method}& \multicolumn{2}{|l|}{GID-DF (HQ)} & \multicolumn{2}{l|}{GID-DF (LQ)} & \multicolumn{2}{l|}{GID-F2F (HQ)} & \multicolumn{2}{l}{GID-F2F (LQ)} \\
        \cline{2-9}
    & ACC            & AUC           & ACC            & AUC           & ACC            & AUC            & ACC            & AUC            \\
    \hline
    EN-B4         & 82.40           & 91.11         & 67.60           & 75.30          & 63.32          & 80.1           & 61.41           & 67.40           \\
    LTW              & 85.60           & 92.70          & 69.15          & 75.60          & 65.60           & 80.20           & 65.70           & 72.40           \\

    DCL             & 87.70           & 94.9         & 75.90          & 83.82         & 68.40          & 82.93          & 67.85          & 75.07    \\
    
    Ours             & \textbf{92.47}           & \textbf{97.42}         & \textbf{82.14}          & \textbf{90.72}         & \textbf{69.12}          & \textbf{84.27}          & \textbf{71.94}          & \textbf{78.60}    \\
    \hline     
    UIA-ViT*             & 90.40           & 96.70         &  -         &     -     & 86.40          & 94.20     & -          & -    \\
        
    Ours*             & \textbf{94.47}           & \textbf{97.42}         & \textbf{85.33}          & \textbf{91.34}         & \textbf{87.93}          & \textbf{96.33}          & \textbf{73.22}          & \textbf{81.25}    \\
    \hline     

    \end{tabular}
    }
    }
    \caption{Performance on multi-source manipulation evaluation, the protocols and results are from \cite{sun2021domain}. GID-DF means traning on the other three manipulated methods of FF++ and test on deepfakes class. The same for the others.}
    \label{table:4}
    \end{table*}

\noindent\textbf{Impact of the different training strategy.}
In Tab.~\ref{table:ab3}, we conduct several training strategies to demonstrate the effectiveness of our C2F. Specifically, Coarse-Only means using a single coarse-grained label to train the encoders; Fine-Pretrain represents that we first leverage fine-grained multimodal learning framework as the pretrain model to learn the forgery-related semantic information, then we adapt with the coarse-grained multimodal learning to obtain binary classification ability. The Fine-and-Fine mechanism uses the PFIG to relabel the original dataset and leverage the fine-grained prompt instead of coarse-grained labels. The result demonstrates our Coarse-and-Fine co-training strategy achieves the best generalization compared to others, which shows the importance of coarse and fine-grained coordination. The Fine-and-fine scheme also obtains suboptimal performance demonstrating that too fine granularity will affect the basic classification ability and generalization.

\begin{table}[!t]
    \centering
    \renewcommand\arraystretch{1.2}
    \resizebox{0.9\columnwidth}{!}{
    \begin{tabular}{c|cc|cc}
        \hline
        \multirow{2}*{Method} & \multicolumn{2}{c|}{Celeb-DF}&
        \multicolumn{2}{c}{DFDC-P}  \\
        \cline{2-5}
        & AUC& EER &AUC& EER \\
        \hline
        Coarse-Only&80.79&25.16&81.79&26.13\\
        Fine-Pretrain&81.29&25.49&82.64&25.76\\
        Fine-and-Fine&82.19&24.43&82.95&25.53\\
        Coarse-and-Fine&\textbf{84.80}&\textbf{22.73}&\textbf{84.74}&\textbf{23.43}\\
        \hline
    \end{tabular}
    }
    \vspace{-2pt}
    \caption{Ablation of the different training strategy.}
    \label{table:ab3}
\end{table}


\section{Additional Visualization Results.}

    
\begin{figure}[!t]
    \begin{center}
    
      \includegraphics[width=1\linewidth]{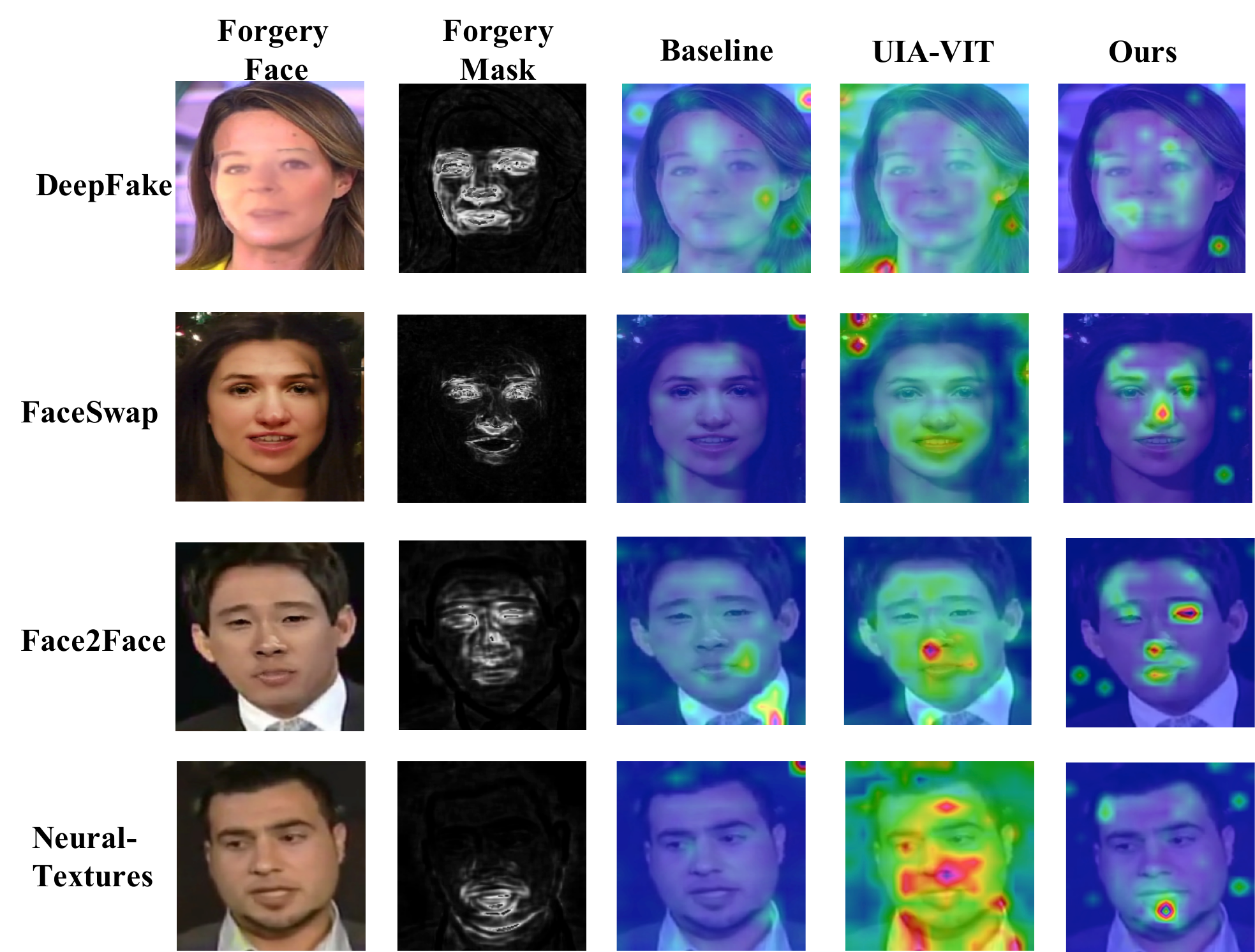}
    \end{center}
      \caption{Visualization of attention heatmap on training dataset (FFpp) of the baseline, UIA-VIT, and our proposed method. Forgery Mask represents the ground truth 
      manipulate mask generated by Eq. 1.
      }
    \label{fig:vis1}
\end{figure}
\begin{figure}[!t]
    \begin{center}
    
      \includegraphics[width=1\linewidth]{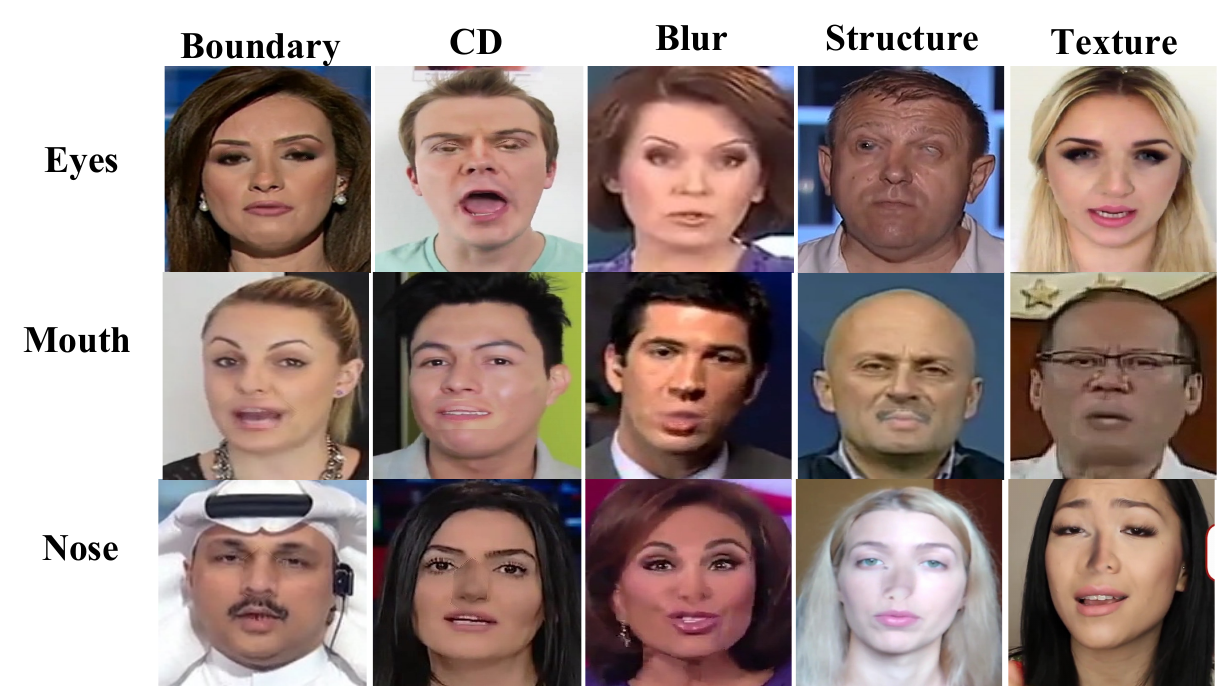}
    \end{center}
      \caption{Visualization of mixed forgery images with different regions and types. Note that there may exist more than one forgery type for each image, here we show the most obvious features. The CD represents color difference; Structure and Texture denote structure abnormal and texture abnormal, respectively.
      }
    \label{fig:vis2}
\end{figure}
\noindent\textbf{Additional visualizations on FFpp dataset.}
To further validate the interpretability of our method, we visualized the attention heatmap of the baseline method and the state-of-the-art method UIA-VIT~\cite{zhuang2022uia} compared with ours on the test set of training data (FFpp HQ), including four methods namely DeepFake, FaceSwap, Face2Face and NeuralTextures, and ground truth mask for comparison. Specifically, we use the official code with the well-trained model on FFpp (HQ) and visualize the attention layer of UIA-VIT.
We can see from Fig.~\ref{fig:vis1} that our method can detect more precise forgery traces compared with baseline and UIA-VIT. For example, our method clearly focuses on the subtle mouth area in the NeuralTextures method, which is consistent with the ground truth, while other methods are misplaced. This demonstrates that fine-grained linguistic information can provide more precise guidance to our detection method.

\noindent\textbf{Visualizations of the mixed forgery image.}
In this section, we provide more examples of the mixed forgery image generated by the PFIG in Fig.~\ref{fig:vis2}. Note that there may exist more than one forgery type for each image, here we show the most obvious features.
From the figure, we observe that each forgery type corresponds to a specific characteristic. For example, the \textit{Blend boundary} has obvious fusion traces in the designated area, and if the shape is deformed, it will be judged as the \textit{Structure abnormal}. 

\begin{figure*}[!t]
    \begin{center}
    
       \includegraphics[width=1\linewidth]{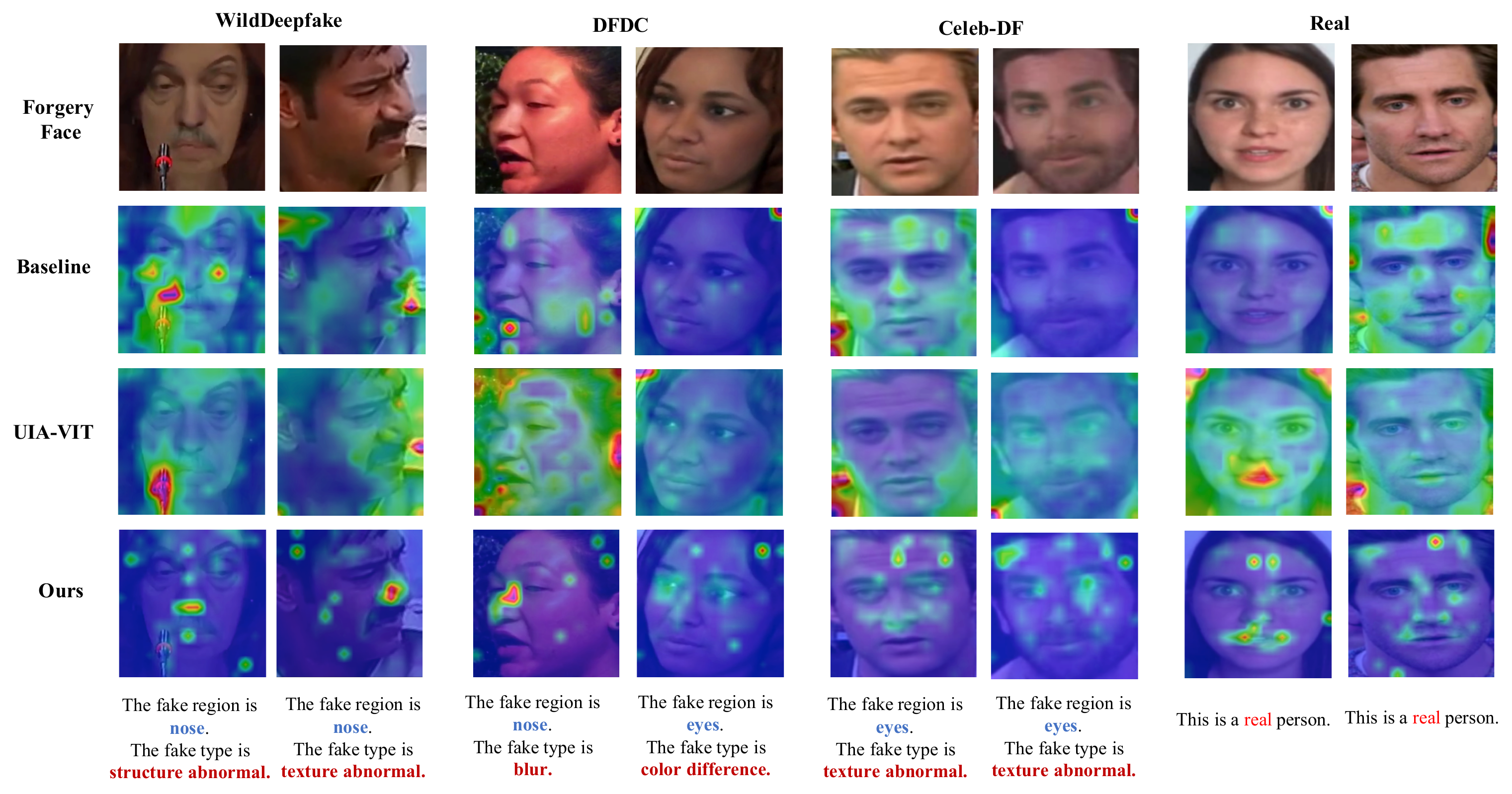}
    \end{center}
       \caption{ Attention heatmap visualization of the baseline, UIA-VIT, and our proposed method on the unseen dataset. The first row represents the original images that did not appear in the training set.
       The last row represents the Top-1 matching prompts of our methods.
       (Best viewed in color.)
       }
    \label{fig:vis1}
\end{figure*}

\noindent\textbf{Additional visualizations on the unseen dataset.}
To further illustrate the superiority of our method intuitively, we have supplemented Figure 5 of the main paper with a more detailed comparison with the state-of-the-art method UIA-VIT~\cite{zhuang2022uia}. Specifically, we use the official code with the well-trained model on FFpp (HQ) and visualize the attention layer of UIA-VIT. The visualization results are shown in Fig.~\ref{fig:vis1}. We can observe that compared with baseline and UIA-VIT, our method can better capture forged features, such as in the first column of images, where both baseline method and UIA-VIT are distracted by the interfering microphone, while our method can focus on the fake eyes and nose of the face. Moreover, for some high-quality forgery faces, such as in the fourth column, our method can highlight more discriminative features (eye color differences). Furthermore, the fine-grained cues provided by our method can also enhance the interpretability of the model, which is beneficial for applying the model in various scenarios.
\section{Details of the testing period.}
During testing, all text features corresponding to coarse-grained and fine-grained prompts can be extracted in advance, and the image feature of the test face is matched with the most similar text feature by cosine similarity to obtain the binary label as well as the text description. The test period can be divided into coarse-grained classification and fine-grained matching. The former aims to identify the real or fake of the input face, while the latter output the sentence-level description in terms of forgery region and types. All the quantization results use the coarse-grained classification.

\end{document}